# Mixat: A Data Set of Bilingual Emirati-English Speech

**Maryam Al Ali, Hanan Aldarmaki**

Mohamed Bin Zayed University of Artificial Intelligence
maryam.alali@mbzuai.ac.ae, hanan.aldarmaki@mbzuai.ac.ae

**Abstract**

This paper introduces Mixat: a dataset of Emirati speech code-mixed with English. Mixat was developed to address the shortcomings of current speech recognition resources when applied to Emirati speech, and in particular, to bilingual Emirati speakers who often mix and switch between their local dialect and English. The data set consists of 15 hours of speech derived from two public podcasts featuring native Emirati speakers, one of which is in the form of conversations between the host and a guest. Therefore, the collection contains examples of Emirati-English code-switching in both formal and natural conversational contexts. In this paper, we describe the process of data collection and annotation, and describe some of the features and statistics of the resulting data set. In addition, we evaluate the performance of pre-trained Arabic and multi-lingual ASR systems on our dataset, demonstrating the shortcomings of existing models on this low-resource dialectal Arabic, and the additional challenge of recognizing code-switching in ASR. The dataset will be made publicly available for research use.

## 1. Introduction

Code-switching (CS), or code-mixing[1], refer to the linguistic behavior of alternating between languages within a conversation or an utterance, which is common in multi-cultural, multi-lingual communities. Code-switching can be sub-categorized as inter-sentential CS (alternating at sentence boundaries), intra-sentential CS (alternating within the same sentence), and even intra-word CS, where languages are mixed within a single word. In this discussion, we use the term code-switching to refer to both inter- and intra-sentential CS, with a particular focus on intra-sentential CS as it is generally more difficult to process using current speech and language technologies.

In the United Arab Emirates (UAE), where Arabic is the primary local language and English is a widely spoken second language, code-switching and code-mixing have become observable and significant aspects of daily communication (Siemund et al., 2021). This is especially true among the younger Emirati population, who frequently engage in code-switching between their native Emirati dialect and English. Several factors contribute to this linguistic phenomenon, including the UAE's diverse expatriate communities that outnumber the native Emirati population, educational systems that promote bilingualism, and the global influence of English as a lingua franca. Studies, such as the one by Kaddoura and Kaddour (2019), highlight the prevalence of code-switching among Emirati youth, underscoring its importance for understanding the linguistic culture of the Emirati population. More generally, the Emirati dialect refers to the dialectal varieties spoken by the native Emirati population, which vary by region to some extent, but are mutually intelligible. Emirati Arabic bears some similarities to dialects from surrounding countries such as Saudi Arabia, Oman, and Qatar, but has its own distinctive characteristics. Currently, speech and language resources that target the Emirati dialects in particular are scarce, and current ASR models trained on other varieties of Arabic[2] do not generalize well to Emirati speech (see section 4 for concrete results).

To study and represent spoken language in the UAE, we need data sets that document the language actually spoken by Emirati people. To that end, and for the purpose of studying intra-sentential code-switching in Emirati Arabic, we collected and annotated a dataset from two podcasts by bilingual Emirati speakers, which represent a common way of speaking by a wide segment of young Emiratis. The resulting data set consists of approximately 15 hours of speech, complete with corresponding transcriptions in Arabic and latin script for clear identification of code-switching points. The speech has been segmented into 5,316 utterances, 1,947 of which include code-switching, while the rest are monolingual Emirati Arabic or English. The following sections describe the data collection and annotation process, data set statistics, and ASR results using existing large pre-trained ASR models: Whisper (Radford et al., 2022), MMS (Pratap et al., 2023), and ArTST (Toyin et al., 2023). We summarize related work in section 5.

---

[1] The terms code-switching and code-mixing are often used interchangeably, but in some fields may refer to related but distinct phenomena. In this paper, we use the terms interchangeably.

[2] Current large data sets consist mostly of MSA, Egyptian, and Saudi Arabic.

## 2. Data Set Construction

For the construction of our dataset, we sourced audio content from online podcasts produced by native Emirati speakers : 'The Direction' podcast[3], and 'Think With Hessa' podcast [4]. These podcasts span a diverse array of topics, such as sports, money and finance, science and technology, and health, and were selected because the hosts often code-switch. With permission obtained from the hosts, we extracted the audio from 14 episodes in Podcast 1, and 14 episodes in podcast 2. For the rest of the paper, the two previously mentioned podcasts will be referred to as 'part 1' and 'part 2' respectively. Part 1 is in the form on conversations between the host and a guest, while part 2 is a structured monologue by a single speaker.

After extracting the content, we split the audio roughly at utterance boundaries, and outsourced the initial round of annotation and validation, which was conducted by Arabic, but non-Emirati, speakers. English speech was transcribed in latin script with the standard English spelling, whereas Arabic speech was transcribed with the Arabic alphabet. This provides a clear separation of the two languages and code-switching points. The second round of validation was conducted by an Emirati speaker to ensure that the annotations reflect conventional Emirati writing patterns[5].

Once the data was fully annotated, we separated the monolingual and CS segments to compute the following statistics.

## 3. Data Set Statistics

The resulting `Mixat`[6] Data set consists of approximately 15 hours of audio content. More than two-thirds of the content is derived from Part 1, which is the conversational podcast. The dataset includes segments of monolingual speech in addition to code-switched speech. While the primary focus of the dataset is the Emirati and Emirati-English code-switched content, there is also a small portion of English-only segments, which we maintained for completeness. In total, 1,947 sentences include code-switching, accounting for 36% of the sentences.

---

[3]https://www.youtube.com/channel/UCZbKz4QeFWbfMVE0fSJeuUw
[4]https://open.spotify.com/show/3yEonEQO8Jfu4plB6B78HE
[5]Generally speaking, Arabic dialects do not have standard writing systems, and people from different regions have somewhat different conventions.
[6]`Mixat` is a code-mixed word that translates into "mixes"; an example of code-mixing in Emirati and other Arabic dialects.

| Mixat - Part 1 | |
|---|---|
| # Sentences | 3728 |
| # Monolingual Arabic Sentences | 2371 |
| # Monolingual English Sentences | 100 |
| # CS Sentences | 1257 |
| Average CMI of CS sentences | 0.12 |
| Mixat - Part 2 | |
| # Sentences | 1588 |
| # Monolingual Arabic Sentences | 895 |
| # Monolingual English Sentences | 3 |
| # CS Sentences | 690 |
| Average CMI of CS sentences | 0.09 |
| Total | |
| Duration (in hours) | 14.9 |
| # Sentences | 5316 |
| # Monolingual Arabic Sentences | 3266 |
| # Monolingual English Sentences | 103 |
| # CS Sentences | 1947 |
| Average CMI of CS sentences | 0.11 |

Table 1: Mixat Dataset Statistics. Part 1 and Part 2 correspond to the two podcasts used, as described in section 2.

Additionally, we calculated the average code mixing index (CMI) of the CS portion of the dataset using the following formula modified from (Chowdhury et al., 2020):

$$CMI^i = w_N \left( \frac{min(N_a^i, N_e^i)}{N^i} \right) + w_\alpha \frac{\alpha^i}{N^i} \quad (1)$$

where $N^i$ is the total number of words in utterance $i$, $N_A^i$ and $N_E^i$ are the total number of Arabic and English words in utterance $i$, respectively, and $\alpha^i$ is the number of code switching points in the same utterance. We use equal weights for the two parts: $w_N = w_\alpha = 0.5$. We report the dataset CMI by averaging the CMIs of all utterances.

Table 1 summarizes the statistics of the whole `Mixat` dataset, as well as individual statistics for parts 1 and 2. Figure 1 shows the distribution of utterance lengths in the two parts of the dataset. The distribution illustrates the different nature of the two parts: Part 1 is derived from a conversational podcast, which is characterized by frequent short utterances, including many one-word utterances that are commonly used in conversations (e.g. 'ok', 'right'). On the other hand, part 2 has a rough normal distribution, reflecting its more formal and structured nature.

## 4. ASR Performance on Mixat

In this section, we report the results on existing Arabic and multi-lingual ASR models that presumably include Arabic as one of their languages. In partic-ular, we evaluate the performance of the following models: Whisper, MMS, and ArTST.

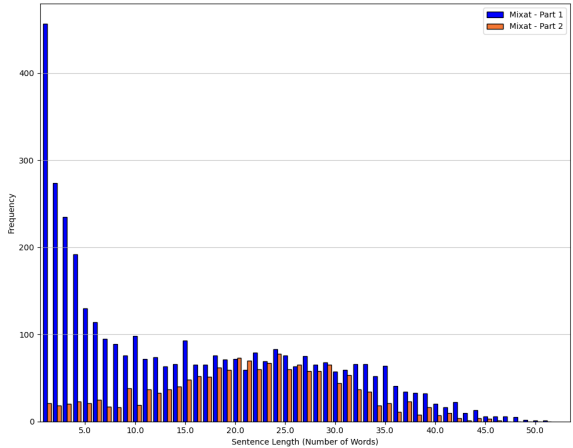

Figure 1: Frequency distribution of sentence lengths in the `Mixat` dataset, measured as the number of words per sentence, with comparisons between `Mixat` Part 1 and Part 2.

Whisper ([Radford et al., 2022](#)) is a multi-task speech-to-text system trained in a supervised manner across many languages and tasks, including speech transcription and translation. Whisper can be used off-the-shelf by providing the language id (e.g. `arabic`) and the task (e.g. `transcribe`) for inference.

The Massively Multilingual Speech (MMS) is another multilingual speech-to-text technology spanning thousands of languages ([Pratap et al., 2023](#)). Similar to Whisper, MMS was pre-trained in a supervised manner ASR across different languages, and the language id can be specified for inference. They use language adapters to optimize the model for different languages.

ArTST ([Toyin et al., 2023](#)) is a pre-trained Arabic text and speech transformer, designed with a focus on the Arabic language, and was pre-trained on a thousand hours of Modern Standard Arabic. Unlike Whisper and MMS, ArTST is not a multilingual model, and is not likely to recognize English, but it has been shown to achieve state-of-the-art performance on Arabic ASR and other speech classification tasks, and was show to have some dialectal coverage.

We selected this combination of multi-lingual and monolingual models to illustrate the performance current state-of-the-art ASR systems on our dataset, illustrating the unique challenges of this low-resource variety.

### 4.1. Evaluation and Results

The overall ASR results are summarized in Table [2], which presents the WER and CER of each model in each part of the dataset. Overall, none of the models provide satisfactory transcriptions for this dataset, rendering them unusable for this task. The WER results are so high that it makes little sense to compare them across model, but the results show that monolingual ArTST is slightly better than the multilingual models. This is likely due to the fact that the majortiy of the dataset is monolingual.

| System | Mixat Segment | WER (%) | CER (%) |
|---|---|---|---|
| Whisper | Part 1 | 204.88 | 233.60 |
|  | Part 2 | 83.20 | 54.12 |
|  | All | 168.52 | 179.97 |
| MMS | Part 1 | 182.6 | 180.0 |
|  | Part 2 | **68.73** | **28.99** |
|  | All | 147.2 | 133.0 |
| ArTST | Part 1 | **118.4** | **115.4** |
|  | Part 2 | 98.9 | 92.1 |
|  | All | **112.2** | **108.0** |

Table 2: Performance of ASR systems on each part of the `Mixat` data set. "All" refers to the combined dataset of Part 1 and Part 2.

We do the evaluation separately for each language and for the CS utterances and show the results in Table 3. For MMS, we used `MMS-1b-all`; for Whisper, we used the `medium` variant, and we used the target language id for the monolingual sentences (e.g. `english` for the English utterancs), and `arabic` for the CS utterances. Whisper's performance is decent for English, and MMS shows improved performance as well; the multilingual models underperformed mainly on the Emirati segments of the dataset, resulting in ∼200% WER. In contrast, these mutlilingual models can recognize Classical Arabic and Modern Standard Arabic, as reported in ([Toyin et al., 2023](#)). ArTST achieved the lowest performance on the Emriati segments of the data, but the WER is still above 100%, showing that transfer from MSA to Emirati Arabic is still challenging.

| System | Language Segment | WER (% | CER (%) |
|---|---|---|---|
| Whisper | Arabic | 195.98 | 255.34 |
|  | English | **12.06** | **11.15** |
|  | Code-Switching | 121.78 | 97.67 |
| MMS | Arabic | 188.1 | 190.7 |
|  | English | 72.75 | 44.90 |
|  | Code-Switching | **90.37** | **52.44** |
| ArTST | Arabic | **119.0** | **118.7** |
|  | English | 341.6 | 401.2 |
|  | Code-Switching | 95.91 | 83.95 |

Table 3: System Performance on monolingual Arabic, monolingual English, and Code-Switched segments of the `Mixat` data set.

| Language | System | Transcription |
|---|---|---|
| Arabic | Reference | إنت ذكرت نقطه اللي برد أنا مره ثانيه بذكرها لإنه موضوع تفويض الصلاحيات وايد مهم |
| | Whisper | أنت ذكرت نقطة التي سأعود لها مرة أخرى لأنها موضوع تفويض الصلاحيات وهي مهمة |
| | MMS | نت ذكارة نقطي لبردا مرثانية باذكرها لأن موضوع تفويض الصلاحيات وهيت مهم |
| | ArTST | ذكرت نقطة اللي بردنا مرارا وتكرارا |
| English | Reference | Life, you will always struggle in life. there are always struggles. |
| | Whisper | Life, you will always struggle in life. There are always struggles. |
| | MMS | life, you will always struggle in life there are always struggles |
| | ArTST | الله الرحمن الرحيم |
| CS | Reference | لكن حان الوقت [ok] إن أي حد يسمع هذا ال [podcast] على الأقل إنه يعني يشك في هذا الكلام. ويشك في هذا المنطق. |
| | Whisper | لكن حان الوقت، أن أي أحد يسمع هذا البرتكاست على الأقل أن يشك في هذا الكلام ويشك في هذا المنطق |
| | MMS | لكن حانا الوقت نيحد اسمع هذا البادكاست على الأقل شف هذا الكلاميشكف هذا المنطق |
| | ArTST | لكن حان الوقت |

Table 4: Examples of Arabic, English, and CS reference transcriptions and ASR hypotheses generated by Whisper, MMS, and ArTST.

### 4.2. Examples

In this section, we provide some examples of transcriptions generated by each model compared to the ground truth to understand their shortcomings. As shown in Table 4, Whisper translates the sentences into MSA, and the translations are often correct. This shows that Whisper in fact recognizes the dialect and the code-switching, but performs the wrong task; this could be an unforseen side effect of the multi-task pre-training, but it shows that there is more potential in this model compared to the alternatives. MMS outputs seem to correspond better to the spoken content, but it is generally ill-formed, possibly due to the lexical shift in this dialect. It also performs worse than Whisper on the English parts, which could be a result of accent differences. ArTST, as a monolingual Arabic model, cannot recognize English words except for frequent short words such as 'ok'. It also produces outputs that are relatively short compared to the input, resulting in many deletions.

### 5. Related Work

In this section, we review code-switching datasets for other variants of dialectal Arabic. ESCWA.CS corpus [7] offers 2.8 hours of dialogue from United Nations sessions including intersentential code-switching between Arabic, English, and French. It offers a resource for studying formal multilingual communication within West Asian UN discussions, typically as a test set only due to its small size. Other data sets that cover dialectal Arabic-French code-swtiching are the Algerian Arabic-French (Amazouz et al., 2018) and Maghrebian Arabic-French (Amazouz et al., 2016) datasets. QASR.CS (Mubarak et al., 2021) Originates from Aljazeera's content include Arabic-French and Arabic-English code-switching. With a Code-Mixing Index (CMI) of 30.5, it illustrates the frequent mixing of languages in media settings. The Egyptian Arabic-English code-switching corpus (Hamed et al., 2020), known as ArzEn corpus contains 12 hours of spontaneous speech collected from 38 bilingual interviews. This dataset is collected from informal bilingual communication among Egyptian university students and employees, serving as a crucial resource for both ASR development and sociolinguistic studies. The Egyptian Arabic-English data set(Hamed et al., 2018) comprises 6 hours of technical domain interviews with informal Egyptian Arabic-English code-switching. Although transcriptions are available for only two-thirds of the content, it remains a valuable asset for understanding technical discourse in a bilingual Egyptian context. A Saudi Arabic-English code-switching dataset is described in (Ismail, 2015), with 89 minutes of transcribed conversations from informal gatherings.

### 6. Conclusion

This paper describes the construction and properties of a newly developed dataset for Emriati-English code-switching, named `Mixat`. It also describes the performance of large pre-trained mulitlingual and monolingual ASR systems on this data set, demonstrating the present difficulties of recognizing spoken Arabic in its low-resource varieties. The data set will be made available for research[8].

### 7. Acknowledgements

We thank Mr. Mohammad Al Awadhi, host of 'The Direction' podcast, and Ms. Hessa Alsuwaidi, host of 'Think With Hessa' podcast for allowing us to use their content for creating a dataset to support academic research on Emirati speech.

---

[7] https://arabicspeech.org/resources/escwacs

[8] github.com/mbzuai-nlp/mixat